\pdfoutput=1

\documentclass[11pt]{article}

\usepackage[final]{acl}

\usepackage{times}
\usepackage{latexsym}

\usepackage[T1]{fontenc}

\usepackage[utf8]{inputenc}

\usepackage{microtype}

\usepackage{inconsolata}
\usepackage{times}
\usepackage{latexsym}
\usepackage{color,xcolor,colortbl}
\definecolor{firered}{RGB}{252,22,17}
\definecolor{iceblue}{RGB}{33,102,240}
\definecolor{router}{RGB}{0, 128, 0}
\definecolor{shared_expert}{RGB}{191, 144, 0}
\definecolor{router_expert}{RGB}{104, 46, 205}
\usepackage{amsfonts,amssymb}
\usepackage{subfigure}
\usepackage{amsthm}
\usepackage{graphicx}

\usepackage{float}
\usepackage{amsmath}
\usepackage{pifont}
\usepackage{bbm}
\usepackage{enumitem}
\usepackage{booktabs}
\usepackage{multirow}
\usepackage{CJKutf8}
\usepackage{xspace}
\usepackage{multirow}  
\usepackage{makecell}  
\usepackage{multicol}
\usepackage{diagbox}
\usepackage{xcolor}
\usepackage[most]{tcolorbox} 
\newtcolorbox{graybox}{
  colback=gray!10,colframe=gray!35,boxrule=0.2pt,arc=1pt,
  left=4pt,right=4pt,top=2pt,bottom=2pt,breakable
}

\usepackage[T1]{fontenc}

\usepackage[utf8]{inputenc}

\usepackage{microtype}
\usepackage{adjustbox}
\usepackage{url}

\newcommand*{\Scale}[2][4]{\scalebox{#1}{$#2$}}

\usepackage{inconsolata}

\usepackage{graphicx}

\definecolor{fired}{RGB}{222,82,57}
\definecolor{fourier_green}{RGB}{0, 128, 0}
\definecolor{visual_yellow}{RGB}{191, 144, 0}
\definecolor{iceblue}{RGB}{33,102,200}
\definecolor{class}{HTML}{bb9726}
\definecolor{prompt}{HTML}{2978b5}
\definecolor{input}{HTML}{53b245}
\definecolor{mygray}{gray}{.9}

\newcommand{\ie}{\textit{i}.\textit{e}.}
\newcommand{\eg}{\textit{e}.\textit{g}.}

\newcommand{\our}{\textsc{MEPT}\xspace}

\title{MEPT: Mixture of Expert Prompt Tuning as a Manifold Mapper}

\author{%
\thead{
  \normalsize{Runjia Zeng$^{\textbf{1}}$~\hspace{5pt}
  Guangyan Sun$^{\textbf{2}}$~\hspace{5pt}
  Qifan Wang$^{\textbf{3}}$~\hspace{5pt}
  Tong Geng$^{\textbf{2,4}}$~\hspace{5pt} 
  Sohail Dianat$^{\textbf{1}}$~\hspace{5pt} }\\\normalsize{
  Xiaotian Han$^{\textbf{5}}$~\hspace{5pt} 
  Raghuveer Rao$^{\textbf{6}}$~\hspace{5pt}
  Xueling Zhang$^{\textbf{1}}$~\hspace{5pt}
  Cheng Han$^{\textbf{7}}$~\hspace{5pt}
  Lifu Huang$^{\textbf{8}}$~\hspace{5pt}
  Dongfang Liu$^{\textbf{1} \dagger}$~\hspace{5pt}}}\vspace{0.1in}\\ \normalsize{
$^{1}$Rochester Institute of Technology ~\hspace{5pt} $^{2}$University of Rochester ~\hspace{5pt} $^{3}$Meta AI ~\hspace{5pt} $^{4}$Rice University}\\
\normalsize{$^{5}$Case Western Reserve University ~\hspace{5pt} $^{6}$U.S. DEVCOM Army Research Laboratory}\\
\normalsize{$^{7}$University of Missouri - Kansas City ~\hspace{5pt} $^{8}$UC Davis ~\hspace{5pt} $^{\dagger}$Corresponding author}
}

\begin{document}
\maketitle

\begin{abstract}

Considering deep neural networks as manifold mappers, the \textit{pretrain-then-fine-tune} paradigm can be interpreted as a two-stage process: pretrain establishes a broad knowledge base, and fine-tune adjusts the model parameters to activate specific neural pathways to align with the target manifold. Although prior fine-tuning approaches demonstrate success, their rigid parameter space limits their ability to dynamically activate appropriate neural pathways, rendering them ill-equipped to adapt flexibly to the diverse and evolving data distributions. In light of this view, we propose a novel approach, Mixture of Expert Prompt Tuning \textbf{(MEPT)}, as an effective and efficient manifold-mapping framework. \our leverages the Mixture of Experts architecture by integrating multiple prompt experts to adaptively learn diverse and non-stationary data distributions. Empirical evaluations demonstrate that \our outperforms several state-of-the-art parameter efficient baselines on SuperGLUE, achieving notable improvements in mean accuracy (\eg, 1.94\%) while significantly reducing activated prompts by 79.25\%. The effectiveness of \our is further supported by theoretical insights from manifold learning and validated through neural activation pathway visualization results. Our code is avaliable at \href{https://runjia.tech/emnlp_mept/}{runjia.tech/emnlp\_mept}.

\end{abstract}

\section{Introduction}
\vspace{-1mm}

\begin{figure}[h!]        
    \centering
    
    \includegraphics[width=0.5\textwidth]{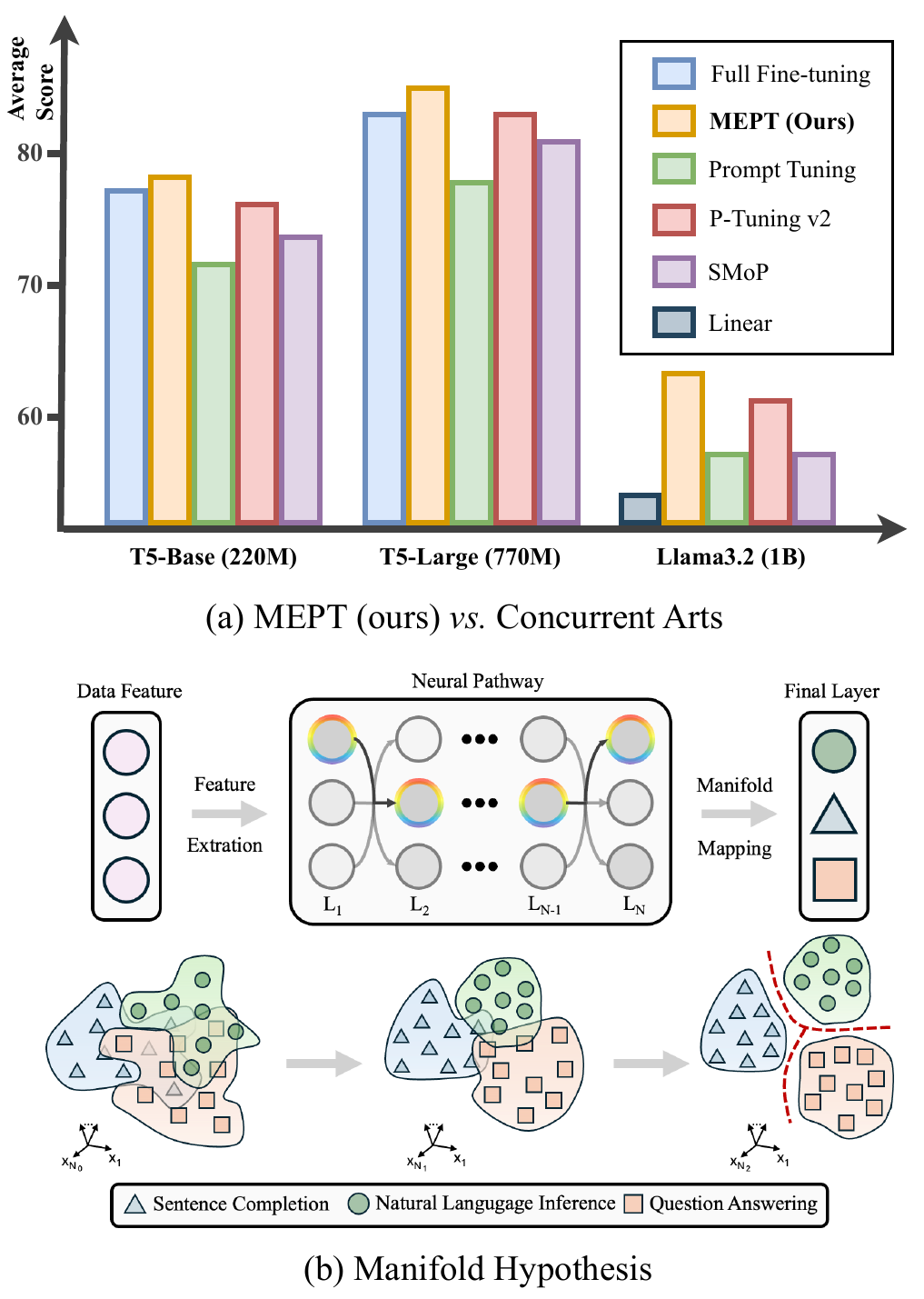}
    \vspace{-8mm}
    \caption{\textbf{MEPT (ours) $vs$ concurrent arts under the manifold learning perspective.} (a) Average accuracy performance comparison with vanilla prompt tuning methods across varying size model structures. (b) Illustration of the manifold hypothesis, where different neural activation paths alter task manifold geometry (blue: ‘Sentence completion,’ green: ‘Natural language inference,’ orange: ‘Question answering’), transforming non-linearly separable manifolds into separable ones in the final layer (separated with red dotted lines).}
    \vspace{-9mm}
    \label{fig1}
\end{figure}

Deep Neural Networks (DNNs) \cite{hinton2006fast} are designed to learn the intrinsic structure of data by mapping inseparable high-dimensional data distributions (\ie, the left tangled feature representation in the Fig.~\ref{fig1} (b)) onto low-dimensional manifolds (\ie, the right linearly separable manifolds in the Fig.~\ref{fig1} (b)). This procedure can be conceptualized as manifold mapping~\cite{cohen2020separability, zhu2018ldmnet}. Through this mapping, DNNs effectively match the data to its most appropriate manifold by eliciting a set of neural responses (\ie, neural pathway highlighted by rainbow ring in the middle of Fig.~\ref{fig1} (b)) that align their learned representations with the intrinsic geometric structure of the data.
This capacity to learn and represent these manifolds is central to the success of DNNs in tasks ranging from image recognition~\cite{hinton2006fast, cohen2020separability} to language modeling~\cite{peters-etal-2018-deep, vaswani2017attention}.

The prevailing training paradigm for large language models (LLMs) is under the \textit{pretrain-then-finetune}~\cite{dodge2020fine, PE1, PE2, hu2021lora} paradigm, which can also be reinterpreted through the lens of manifold learning. In this perspective, pretrain serves in the role to establish a broad knowledge base by exposing the model to diverse data distributions, enabling it to capture a general representation of various manifolds. Finetune then modifies the model's parameters to activate unique neuron pathways that align with the target task’s data manifold. 

Traditional full fine-tuning is parameter-intensive and does not efficiently utilize the general knowledge acquired during pre-training. As LLMs continue to scale in size, researchers have introduced the \textit{parameter-efficient fine-tuning} (PEFT) paradigm to address these limitations.
Though highly successful, current PEFT techniques (\eg, partial tuning~\cite{chen2021empirical,jia2021exploring,mahajan2018exploring}, low-rank adaptation (LoRA)~\cite{jie2023revisiting}, prompt tuning~\cite{ju2022prompting, dong2023efficient, yan2023prompt, zang2022unified}) are constrained by their reliance on fixed parameter spaces: They fall short in dynamically activating the appropriate neuron pathways necessary for effectively handling real-world data, which often entails complex, overlapping, or unknown manifolds. As a result, they are ill-equipped to flexibly adapt to the diverse and evolving data distributions in practice.

In light of this view, we propose \textbf{M}ixture of \textbf{E}xpert (MoE) \textbf{P}rompt \textbf{T}uning (\textbf{\our}).
By leveraging the principles of MoE to activate specialized pathways within the LLMs, we are able to dynamically adapt the network’s representational capacity with the requirements of the task-specific data manifolds. \our exhibits several compelling advantages: \ding{182}~\textit{\textbf{Generality.}} By employing a modular and adaptive MoE design, \our facilitates the handling of diverse and non-stationary data manifold mapping, significantly enhancing the model’s ability to represent and process the complex structures inherent in real-world datasets (see \S\ref{experiment:main_result} and \S\ref{sec:expert}); \ding{183}~\textit{\textbf{Scalability.}} The sparse MoE architecture specializes in the representation of each prompt expert, enabling a compact and efficient neural pathway activation, which imposes no additional burden when scaling up the number of experts during training and inference (see \S\ref{method:ept}, \S\ref{sec:abla} and Tab.~\ref{overall_comparison}); \ding{184}~\textit{\textbf{Interpretability.}} We find that the effectiveness of our approach is directly supported by theoretical insights from manifold learning and further validated through instinct neural activation pathway visualization results (see \S\ref{sec:expert}), demonstrating the capability for flexible and effective manifold mapping. In contrast, existing PEFT methods lack an intuitive manifold structure, thereby rendering their operation opaque and akin to a black box.

The remainder of the paper is structured as follows:
we begin by theoretically re-examining the vanilla prompt tuning approach in \S\ref{met:pre}, showing that it can be considered as a specialized \our\ version with only one expert. We also introduce the general notations for the mixture of experts. The formal introduction of \our\ is presented in \S\ref{method:ept}. 
Comprehensive evaluation on the SuperGLUE benchmark under various LLMs in \S\ref{experiment:main_result} showcases the superior performance of \our compared to state-of-the-art methods. Further ablation studies in \S\ref{sec:abla} provide strong evidence for the effectiveness and efficiency of our proposed expert prompts. To explore the valuable interpretability offered by explicit manifold mapping, we also visualize the neural pathway activation and final manifold representation in \S\ref{sec:expert}. Our main contributions are summarized as the following:
\begin{itemize}
    \item We unveil the essence of neural network learning through manifold visualization, cross-validating our motivation and effectiveness under manifold learning theory while elucidating the decision-making process.
    \item We further investigate the potential of mixture training using a universal prompt embedding, which imposes higher requirements for prompt generality. It highlights the importance of MoE, providing valuable insights for future PEFT architecture design.
    \item We conduct comprehensive experiments on various tasks and LLM sizes in the SuperGLUE benchmark, demonstrating the effectiveness of \our over several state-of-the-art prompt tuning and full fine-tuning methods.
\end{itemize}

\section{Related works}

\subsection{Parameter-Efficient Fine-Tuning}\label{rw:pt}

Large Language Models (LLMs) push forward the state-of-the-art on natural language processing through their remarkable capabilities in various tasks \citep{devlin2018bert,radford2019language,brown2020language}. However, as the size of LLMs continue to
grow, the computational cost of fine-tuning these models has increasingly become parameter-intensive,
posing significant challenges for practical applications \citep{kaplan2020scalinglawsneurallanguage, han2023e2vpt}. To address this problem, PEFT has emerged as a promising solution that significantly reduces computational and memory requirements while maintaining model performance \citep{houlsby2019parameter, hu2021lora, lester2021power, pfeiffer2021adapterfusion}. 
Three PEFT paradigms have gained particular prominence in recent years. 

\noindent\textbf{Adapter-based} approaches introduce small, learnable neural networks (\ie, adapters) between the layers of the pre-trained model \citep{houlsby2019parameter, pfeiffer2020mad}. Adapters typically consist of down-projection and up-projection layers, creating a bottleneck architecture that effectively captures task-specific information while keeping the original model parameters frozen \citep{houlsby2019parameter, PE2, pfeiffer2021adapterfusion, karimi2021compacter}. \\
\noindent\textbf{Low-Rank Adaptation (LoRA)} \citep{hu2021lora} represents a reparameterization approach that decomposes weight updates into low-rank matrices. This method has sparked numerous innovations in dynamic rank selection and efficiency improvements (\eg, DyLoRA \citep{valipour2022dylora}, AdaLoRA \citep{zhang2023adaptive}, LoRA Dropout \citep{lin2024lora}, LoRA+ \citep{hayou2024lora}). \\
\noindent\textbf{Prompt Tuning} 
\citep{lester2021power, PE2} introduces learnable continuous prompts to the input sequence while keeping the pre-trained model frozen. Unlike discrete prompts that use natural language tokens, the continuous prompts are directly optimized during fine-tuning \citep{liu2021prompt, vu2021spot, choi-etal-2023-smop, li2025pt}. 
Different from prior arts that fix tunable parameter representations, \our activates different experts for each manifold mapping, enabling more dynamic adaptation.

\subsection{Mixture of Experts}
Mixture of Experts (MoE) \citep{jacobs1991adaptive, eigen2014learningfactoredrepresentationsdeep} represents a significant advancement in model architecture design.
In MoE architectures, the model consists of multiple specialized sub-networks ($a.k.a.$ experts) that are selectively activated through a gating mechanism \citep{shazeer2017, fedus2022switchtransformersscalingtrillion}. This design enables parameter scaling without proportional increases in computational cost, making it particularly valuable for large pre-trained models. Recent advancements have also extended MoE's applicability to three PEFT paradigms \citep{wang2022adamix, wu2024mixture, choi-etal-2023-smop}.
The promising advancements of MoE inspire our design, aligning with our motivation to establish a dynamic and explainable fine-tuning pipeline.
\begin{figure*}[!ht]
  \centering
  \includegraphics[width=0.95\linewidth]{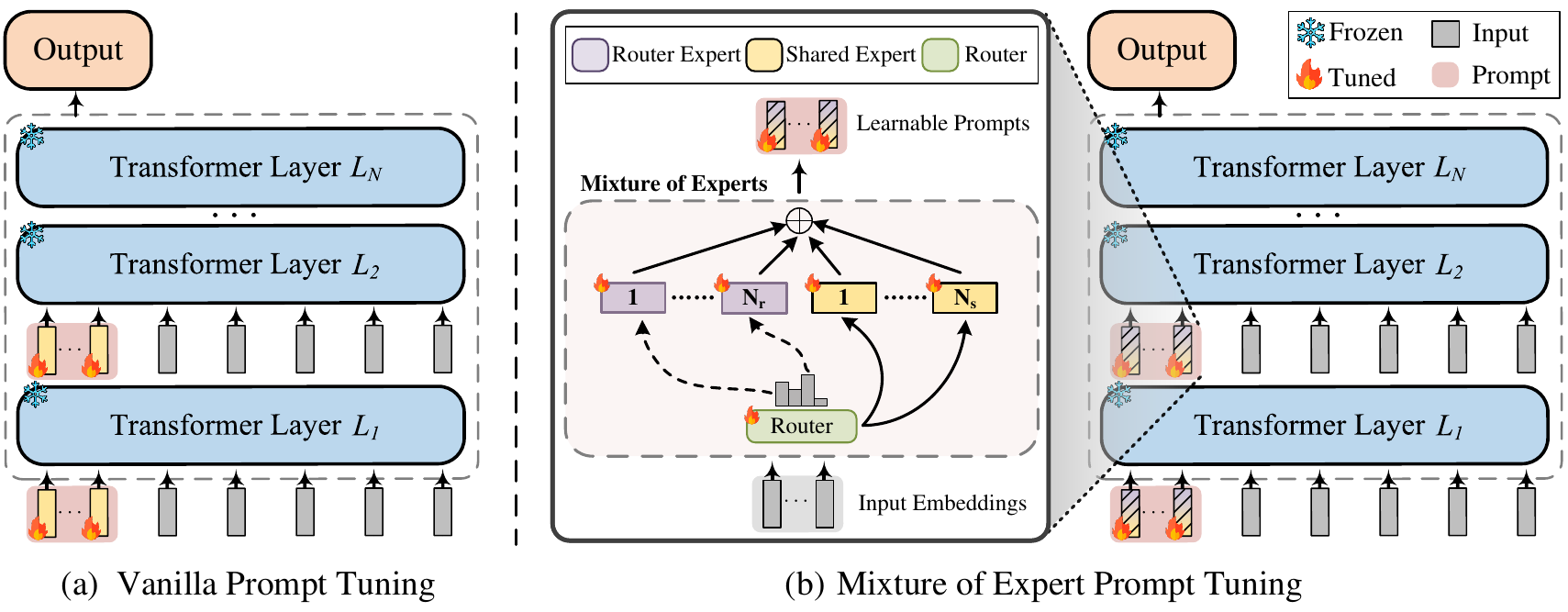}
  \vspace{-4mm}
  \caption{\textbf{Overview of PT $vs.$ MEPT (ours) frameworks.} (a) Vanilla Prompt Tuning. (c) The overall architecture of our proposed \our (see \S\ref{method:ept}), including \textcolor{router_expert}{router expert}, \textcolor{shared_expert}{shared expert} and \textcolor{router}{router} in each layer. } 
  \label{fig2}
  \vspace{-5mm}
\end{figure*}
\section{Methodology}

\subsection{Preliminaries}
\label{met:pre}

\textbf{Prompt Tuning.} In traditional prompt tuning methods \cite{lester2021power, PrefixTuning}, input prompts consist of a set of $d$-dimensional embedding vectors, where the dimensionality matches that of the text tokens. These prompts are inserted at the beginning of the input sequence in each Transformer encoder layer and interact with all the text tokens. They facilitate the learning of task-specific embeddings, effectively guiding the model's performance on new tasks.

Formally, these input prompts are defined as $P$ = $\{P^1, P^2, \dots, P^N\}$, where $P^j$ denotes the learnable input prompts in the $j_{th}$ Transformer encoder layer $L_j$, and $N$ is the total number of layers. Then the outputs of encoder layers are represented as:
\[\Scale[0.9]{Z^j = \textcolor{iceblue}{L_j}(\textcolor{firered}{P^j}, \ Z^{j-1}) \ \ \ \ \ j = 1, 2, 3,\dots,N}\]
where $Z^j$ represents the contextual embeddings of the text tokens computed by the $j_{th}$ encoder layer. The different colors indicate \textcolor{firered}{trainable} and \textcolor{iceblue}{frozen} parameters, respectively. $Z^0$ is the text token embeddings initialized from the backbone.

\noindent\textbf{Mixture of Experts.} In Transformer-based LLMs, each MoE layer generally consists of a set of $K$ ``expert networks'' $\{f_{1}, \ldots, f_{K}\}$, alongside a ``gating network'' $\mathcal{G}$. Formally, the gating network $\mathcal{G}$, parameterized by $\mathbf{\Theta}$ and typically consisting of a linear-softmax network, yields the output $\mathcal{G}(\mathbf{x}; \mathbf{\Theta})$. Consequently, the output of the dense MoE layer can be formulated as:
\[\Scale[0.9]{\mathcal{F}(\mathbf{x}; \mathbf{\Theta}) =
\sum_{i=1}^{K}\mathcal{G}(\mathbf{x}; \mathbf{\Theta})_i f_{i}}\]
\[\Scale[0.9]{\mathcal{G}(\mathbf{x}; \mathbf{\Theta})_i = \operatorname{softmax}(g(\mathbf{x}; \mathbf{\Theta}))_i = \frac{\exp(g(\mathbf{x}; \mathbf{\Theta})_i)}{\sum_{j=1}^{K} \exp(g(\mathbf{x}; \mathbf{\Theta})_j)}}\]
where $g(\mathbf{x}; \mathbf{\Theta})$ represents the gating value prior to the softmax operation.

\subsection{\our}\label{method:ept}

In this paper, we present a novel prompt tuning approach, \our, for dynamic and general LLM adaptation (see Fig.~\ref{fig2}(b)).
The model employs a stacked MoE architecture for prompt embeddings, facilitating multi-layer representation learning to achieve universal adaptation. At each layer, prompt representations are constructed by two types of experts: \textit{router experts} and \textit{shared experts}.

\noindent\textbf{Router Experts as Representation Specialists.} 
While conventional prompts are effective in acquiring knowledge about the new task, they do not possess the capability to flexibly map each input to the most suitable lower-dimensional manifold. During fine-tuning on a new task with new data, the word distributions may significantly differ from each other. Consequently, it becomes imperative to enhance the model's capacity for activating different neural pathways for better manifold mapping. This entails enabling better attention among the mixture of expert prompt to flexibly learn new patterns that emerge in the task-specific context. Specifically, the $j$-th layer knowledge is selected from $N_r$ router experts as:
\[\Scale[0.9]{\mathbf{RE^{j}}  = Top(\textcolor{router}{\mathcal{G}(\mathbf{h^{j-1}}; \mathbf{\Theta})_i}})\textcolor{router_expert}{R_i^j}\ \ \ \ \ j = 1, 2, \dots,N\]
where $R^j \in \mathbb{R}^{N_r \times m \times d} $ denote the embeddings of the router expert prompts in the $j$-th transformer encoder/decoder layer, and $N_r$, $m$, and $d$ represent the number of router experts, the length of the prompt, and the dimension of hidden state, respectively. \adjustbox{valign=c}{\includegraphics[scale=0.10]{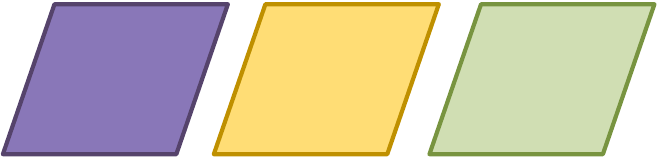}} colors represent router experts, shared experts, and router, respectively. The router, denoted by $\mathcal{G}$, is implemented as a linear network followed by a softmax activation. It is responsible for directing the input $h^0$ (in the first layer) or the hidden state $h^j$ ($j\in [1, N]$) from the preceding layer to the appropriate expert prompts (in subsequent layers). Finally, we sparsely activate only the $i$-th ($i\in [1, N_r]$) prompt expert with the highest routing probability through the Top($\cdot$).

These router experts are explicitly designed to specialize in capturing diverse aspects of input data and task-specific features. By dynamically routing inputs through different experts (\ie, see the pathway visualizations in Fig.~\ref{fig:vis}), the model effectively learns diverse patterns critical for task adaptation (\eg, the superior performance validated in Tab.~\ref{overall_comparison}). Unlike traditional prompt tuning with fixed mappings, \our utilizes specialized prompts tailored to distinct subspaces of the input distribution.

\noindent\textbf{Shared Experts as Knowledge Consolidators.} 
Traditional MoE routing strategies often result in parameter redundancy, as multiple experts may encode overlapping common knowledge \cite{dai2024deepseekmoe} (\ie, vanilla prompt tuning typically requires prompt lengths of 10 to 100 to store the parameters). To address this, shared experts are introduced to centralize and integrate common knowledge across contexts (\ie, \our only activates 10–20 prompts per layer). Formally, the shared expert prompt embeddings in $j$-th layer can be represented as:
\[\Scale[0.9]{\mathbf{SE^{j}}  = \textcolor{shared_expert}{S_1^j} + \cdots + \textcolor{shared_expert}{S^j_{N_s}}}\]
where $S^j \in \mathbb{R}^{ N_s \times m \times d} $ denotes the shared expert prompts in the $j$-th transformer encoder layer, and $N_s$ represent the number of shared experts. Its isolated, router-irrelevant design indicates consistent activation, facilitating the retention of common knowledge (see Appendix \S\ref{appendix:caseshared}) while enhancing the specificity of expert knowledge learning for the router experts. We provide further discussion with DeepSeek \cite{dai2024deepseekmoe} in Appendix~\ref{keydis}.

\noindent\textbf{Training MEPT is to build a manifold mapper.} During training, \our separately updates the router experts $R^j$, shared experts $S^j$ and router $\mathcal{G}$ in each layer. This process enables \our to mine distinct data patterns through the router experts to enable a more precise manifold mapping while concurrently consolidating common knowledge into the shared experts, mitigating parameter redundancy that previously stored in the router prompt. Since the prompts in each layer share the same architectural design, the embeddings of the prompt in the $j$-th layer can be represented as:
\[\Scale[0.9]{P^j =\mathbf{RE^{j}} + \mathbf{SE^{j}}} \]
Training with \our at each layer effectively constructs a manifold mapper, as this process progressively refines the representation space (\ie, facilitate within-class convergence while keeping between-class separation, see detailed analysis in \S\ref{sec:expert}), causing it to ``collapse'' around manifold-relevant information (class-specific prototype hidden vector). This system has two merits:
\begin{itemize}[leftmargin=*]
\vspace{-1mm}
\item \textit{Flexible Mapping:}
\our exhibits an emergent capability to dynamically activate distinct expert prompts tailored to the input, thereby mapping them uniquely onto the corresponding manifolds (see analysis in \S\ref{sec:expert}), bringing consistently superior performance across various tasks, backbones and training schemes (see Tab.~\ref{overall_comparison} and \ref{tab:combined}).

\item \textit{Efficient Representing}: Its intrinsically efficient parameter space leverages a leaner approach (\eg, the parameter count for the prompt-based method is $mh$ (see analysis in \S\ref{method:ept})), enabling more scalable growth compared with the traditional MoE (\ie, the parameter count of the MLP component in a transformer is approximately $8h^2$~\cite{korthikanti2023reducing}, where $h$ denotes the hidden layer dimension).

\end{itemize}

\section{Experiments}

\begin{table*}[t]
\begin{adjustbox}{width=0.75\width,center}
\begin{tabular}{c|c|cccccc>{\columncolor[gray]{0.9}}c}
\toprule
\multirow{2}{*}{\textbf{Method}} & \multirow{2}{*}{\textbf{Para}}  & \textbf{Boolq} & \textbf{CB} & \textbf{COPA} &  \textbf{MRC}  &  \textbf{RTE} & \textbf{WiC}   &\textbf{Average}   \\  
&&Acc &F1/Acc &Acc &F1a  &Acc & Acc &Score\\
\midrule
\midrule
\multicolumn{9}{c}{\textbf{T5-Base} (220M)}\\
\midrule
Fine-Tuning$\dagger$ \cite{ExT5} &100\%   & 82.30 & 91.30 & 60.00 & 79.70  & 84.50  & 69.30 & 77.85 \\
\midrule
Prompt-Tuning$^*$ \cite{lester2021power}&0.06\% & 78.12 & 84.42 & 54.37 & 78.30  & 75.27 & 62.29  & 72.13\\
P-Tuning v2$^*$ \cite{ptuningv2}&0.53\%& \underline{80.81} & \underline{90.23} & 61.28 & \underline{79.83}  & \underline{81.98} & 67.56 & \underline{76.94} \\
XPrompt$^*$ \cite{xprompt}&0.04\%      & 79.67 & 86.72 & 56.95 & 78.57 & 78.29 & 64.31  &  74.09            \\ 
ResPrompt$^*$ \cite{ResidualPrompt}&0.21\% & 79.25 & 85.33 & 58.64 & 78.42  & 77.14 & 62.36  &  73.52   \\
SMoP$\dagger$ \cite{choi-etal-2023-smop}&   0.01\%         & 79.40 &86.42&58.30& 79.60 &77.50&65.20& 74.40 \\
DePT$\dagger$ \cite{shi2023dept}&   -       & 79.30 &-&-& - &79.10&\underline{68.70}& - \\
SuperPos-Prompt$\dagger$ \cite{pmlr-v262-ali-sadraei-javaheri24a}&   -       & 74.00 &80.20&\underline{62.00}& 72.90 &70.40&67.60& 71.18 \\
VFPT \cite{zeng2024visual}&   0.21\%       & 78.38   & 90.92 & 61.76 & 78.73 & 76.90 & 65.36 & 75.34 \\
EPT \cite{lan2024efficient}&   0.06\%       & 79.14   & 90.18 & 56.33 & 73.43 & 78.99 & 67.71 & 74.30 \\
\our (Ours) &    0.13\%        & \bf 81.13 & \bf 90.92 & \bf 62.33  & \bf 80.70 & \bf 83.52 & \underline{69.60} &  \bf 78.03    \\
\midrule
\midrule
\multicolumn{9}{c}{\textbf{T5-Large} (770M)}\\
\midrule
Fine-Tuning \cite{ExT5}&100\%  & 85.75 & 95.26 & 76.00 & 84.41 &  88.05  &72.11 & 83.60    \\ 
\midrule
Prompt-Tuning \cite{lester2021power}&0.04\%& 83.20 & 90.32 & 57.50 & 83.10 & 86.11  &68.74 & 78.16    \\
P-Tuning v2 \cite{ptuningv2}&0.52\% & \underline{85.82} & \underline{95.56} & 77.00 &  84.07 & \underline{89.25}  &71.03   & 83.79     \\
XPrompt$^*$ \cite{xprompt}&0.02\%     & 85.54 & 91.39 & \bf 85.05 & \underline{84.36} & 87.30  
&\bf73.22  & \underline{84.48} \\ 
ResPrompt$^*$ \cite{ResidualPrompt}&0.15\%& 83.51 & 90.64 & \underline{82.79} & 84.02 & 86.97   &71.13  &   83.18    \\
SMoP \cite{choi-etal-2023-smop}&   0.04\%   & 83.45 &92.37&71.00&83.92&87.70&68.60& 81.17 \\
VFPT \cite{zeng2024visual}&   0.18\%       & 83.89 & 93.71 & 75.63 & 83.24 & 88.10 &  71.00&82.56\\
EPT \cite{lan2024efficient}&   0.04\%       & 84.77 & 93.40 & 54.00 & 80.03 &86.33 &  71.79  & 78.39\\
\our (Ours) &    0.12\%       & \bf 85.96 & \bf 97.61& 79.67&\bf85.29&\bf90.01& \underline{73.09}& \bf 85.27    \\
\midrule
\midrule
\multicolumn{9}{c}{\textbf{Llama-3.2} (1B)}\\
\midrule
Linear Head \cite{dubey2024llama}&  3e-4\%       &59.85& 51.69 & 56.33& 48.94&55.23&53.45&54.25 \\
\midrule
Prompt-Tuning \cite{lester2021power}&    0.06\%     & 60.95&61.61&57.67&57.73&55.96&54.70& 56.94 \\
P-Tuning v2 \cite{ptuningv2}&   0.53\%    & \underline{62.48} & \underline{64.29} &\underline{61.00}&\underline{60.34}&\underline{58.12}&\bf60.15&\underline{61.06} \\
SMoP \cite{choi-etal-2023-smop}&   0.04\%        &61.13 &62.50&59.33&57.46&57.40&54.23&57.51 \\
VFPT \cite{zeng2024visual}&   0.17\%       & 62.44 & 61.72 & 59.67 & 58.41 &\bf64.35 & 57.60&60.70 \\
EPT \cite{lan2024efficient}&   0.06\%       &  61.56 &65.22 &56.00 & 60.18 & \underline{63.90}&  59.45&61.05 \\
\our (Ours) &    0.11\%      & \bf64.56&\bf67.86&\bf63.33&\bf62.22&60.29&\underline{59.72}&\bf63.00     \\
\bottomrule
\end{tabular}
\end{adjustbox}
\caption{Performance comparison result (\%) on SuperGLUE development set. `$^\ast$' and $\dagger$ indicates the results reported from \cite{aprompt} or their corresponding paper, respectively. `Para' is the number of trainable parameters. The best results except fine-tuning are in bold with underline representing the second best ones. For tasks with two metrics, the average score is reported. All scores are averaged over 3 runs. Results are statistically significant with respect to all baselines on each PLM (all p-value < 0.005).}
 \label{overall_comparison}
 \vspace{-5mm}
\end{table*}

\subsection{Datasets}
Following previous works on prompt tuning \cite{lester2021power,choi-etal-2023-smop}, we use six NLU tasks from the SuperGLUE benchmark to evaluate the performance of the language model \cite{T5,ExT5}, including BoolQ \cite{Boolq}, CB \cite{CommitmentBank}, COPA \cite{Copa}, MRC \cite{MultiRC}, RTE \cite{RTE}, WiC \cite{WIC} and WSC \cite{WSC}. More details are provided in the Appendix \S\ref{appendix:dataset}.

\subsection{Baselines}
Our model is compared with full fine-tuning, linear adaptation (\ie, only tuning the final linear layer in Llama-3.2 1B) and multiple shallow and deep prompt tuning methods. More PEFT comparisons are provided in the Appendix \S\ref{appendix:peft}.
\vspace{-1mm}

\subsection{Implementation Details}

\our \ is implemented with the Huggingface PEFT library \cite{peft}, which is a unified and extensible toolkit for PEFT research. Our model is trained on NVIDIA A100-40GB GPUs. We translate each SuperGLUE dataset into a text-to-text format following \cite{choi-etal-2023-smop}. Three scales pre-trained models are used: T5-Base, T5-Large and Llama3.2 with 200M, 770M and 1B parameters, respectively. Specifically, we train our prompts for 70 epochs using a batch size ranging from 16 to 32 (\ie, to avoid out-of-memory issues on a single A100 GPU) under different learning rate settings. There are two hyperparameters in our model: the lengthes of prompt and number of router experts. For our method, we linearly search the best prompt length from \{10, 15, 20\} and number of router experts from \{4, 10, 20\}. Notably, we only utilize one single shared expert per layer to improve efficiency. To guarantee reproducibility, our full implementation shall be publicly released upon paper acceptance. See more details in \S\ref{appendix:details}.

\subsection{Main Results}\label{experiment:main_result}
The main performance comparison results are shown in Tab.~\ref{overall_comparison}, leading to three key observations.
\textbf{First,} \our not only outperforms all vanilla prompt tuning baselines but also surpasses full fine-tuning, achieving superior accuracy. On T5-Base, it improves Acc by 0.18\% and 1.09\% over full fine-tuning and prior best models, respectively. This highlights the effectiveness of the prompt expert in enhancing performance. In contrast, existing less effective prompt tuning methods focus solely on updating input prompt tokens, neglecting the importance of activating distinct neural pathways based on the target data manifold. \textbf{Second,} \our generalizes well across different model sizes and architectures while maintaining consistently superior performance. Even on the decoder-only Llama model, which is pretrained with a causal language modeling (CLM) objective, \our outperforms linear adaptation and prior best methods by 8.75\% and 1.94\%, respectively, highlighting its broad applicability. \textbf{Third,} \our benefits from an efficient and specialized prompt representation enabled by its sparse MoE routing design. Compared to traditional deep prompt tuning, P-Tuning v2, which also prepend prompts at each layer, \our reduces trainable parameters by up to 79.25\% (\ie, 0.53\% $vs.$ 0.11\%). Another interesting finding is the performance gap between LLaMA 3.2 and T5-Base on SuperGLUE under prompt tuning, which may be attributed to differences in pretraining objectives and architecture (decoder-only $vs.$ encoder-decoder, see detailed analysis in \S\ref{appendix:details}). This observation is consistent with \cite{hu2024improving}.

\section{Ablation Study}\label{sec:abla}
\noindent\textbf{Prompt Depth.}
We investigate the performance of \our based on the specific layers where prompt experts are deployed, as shown in Tab.~\ref{tab:depth}. The results indicate that incorporating prompt experts at a single layer (\ie, shallow) can also enhances accuracy. Furthermore, extending prompt experts to deeper layers (\ie, default $vs.$ deep) leads to the highest overall performance.

\noindent\textbf{Routing Methods.} 
\begin{table}[t!]
\small
    \centering
    \vspace{-3mm}
    \begin{tabular}{ll| ccccc}
\toprule
\multicolumn{2}{c}{\our} &  CB & COPA  & WiC & MRC \\\midrule
(Default)&1-12 & \bf 90.9& \bf 62.3&\bf 69.6&\bf  80.7 \\\midrule
\multirow{2}{*}{(Deep)}&1-6 & 89.8&62.0&67.0& 80.1  \\
&7-12 & 89.9 & 61.3 &66.1& 80.0 \\\midrule
\multirow{2}{*}{(Shallow)}&1 &86.4  &58.3&65.2&79.6   \\
&12 & 85.9 & 58.0 & 65.1 & 79.8 \\
 \bottomrule
 \end{tabular}
 \vspace{-2mm}
    \caption{Performance comparison with different prompt
positions on CB, COPA, WiC and MRC for T5-Base. 
    }
    \label{tab:depth}
    \vspace{-5mm}
\end{table}
\begin{table}[t!]
\begin{adjustbox}{width=1\columnwidth,center}
    \begin{tabular}{l| ccccc}
\toprule
\multicolumn{1}{c}{Routing Method} &  CB & COPA  & WiC & MRC \\\midrule
\our (Default) & \bf 90.9& \bf 62.3&\bf 69.6&\bf  80.7  \\\midrule
\hspace{0.2cm} -  \hspace{0.1cm}Stochastic & 83.2 & 57.7 & 64.9 & 69.6   \\
\hspace{0.2cm} -  \hspace{0.1cm}Dense& 85.7 & 59.0 & 66.0 & 79.7 \\
\hspace{0.2cm} -  \hspace{0.1cm}Gumbel-Softmax& 88.3 & 61.3 & 65.1 & 80.0 \\
\hspace{0.2cm} -  \hspace{0.1cm}Perturbation& 90.1 & 62.0 & 67.4 & 80.3 \\
\hspace{0.2cm} -  \hspace{0.1cm}w/o Shared Experts& 89.3 & 60.7 & 67.0 & 80.1 \\
\hspace{0.2cm} -  \hspace{0.1cm}Replace Shared Experts & 89.9 & 62.0 & 67.9& 80.2 \\
 \bottomrule
 \end{tabular}\end{adjustbox}
 \vspace{-2mm}
    \caption{Different router variants of \our in T5-Base. 
    }
    \label{tab:router}
    \vspace{-8mm}
\end{table} 
We conduct ablation studies on alternative routing methods in Tab.~\ref{tab:router} to assess the impact of the router, as discussed in \S~\ref{method:ept}. When applying stochastic operations within the router, a dramatic performance decline is observed. This degradation likely arises from the model’s inability to consistently activate the most appropriate experts. Similarly, adding perturbations, enabling dense activation, or incorporating the Gumbel operation in the router also reduces effectiveness, as these modifications constrain the model’s specialization. Lastly, we examine the influence of shared experts by directly removing them from the default version of \our and replacing them with the router experts (comparable parameters with the default), respectively. Notably, without shared experts, the limited prompt embeddings may struggle to capture sufficient information, leading to performance degradation. However, increasing parameter storage helps alleviate this issue. This highlights the critical role of shared experts in efficiently storing and leveraging ``common knowledge.''

\begin{table}[t!]
\small
    \centering
    \begin{tabular}{cc| cccc}
\toprule
\multicolumn{2}{c|}{\textbf{Learning Space}} &  \multirow{2}{*}{CB} & \multirow{2}{*}{COPA}  & \multirow{2}{*}{WiC} & \multirow{2}{*}{MRC} \\
Router&Prompt&&&\\\midrule
\checkmark & \checkmark & \bf 90.9& \bf 62.3&\bf 69.6&\bf  80.7  \\\midrule
&\checkmark &89.1 &61.3&66.3&79.4   \\
\checkmark& & 88.5 &58.0&65.0& 78.3 \\
 \bottomrule
 \end{tabular}
 \vspace{-2mm}
    \caption{Ablative study of router and expert prompt.
    }
    \label{tab:router_prompt}
    \vspace{-3mm}
\end{table}

\noindent\textbf{Variants of \our.} A fundamental difference between \our and other vanilla prompt tuning approaches lies in the integration the MoE into prompt embeddings (see \S\ref{method:ept}). In our standard implementation, the router and prompt are both updated at each training step. To investigate the impact of this design, we analyze the contributions of these components individually in Tab.~\ref{tab:router_prompt}. As seen, the separate updating strategy significantly restricts the dynamic learning of manifold mappings. In contrast, simultaneously updating the router and prompt components exhibits a synergistic effect, leading to substantial performance improvements.

\noindent\textbf{Prompt Length and Number of Experts.}
\begin{table}[t!]
\small
    \centering
    \begin{tabular}{c| ccccc  }
\toprule 
\diagbox{\textit{L}}{\textit{$N_r$}}& 4 & 6& 10 & 14 & 20   \\\midrule
10 & 83.0 & 82.0 & 81.8 & 82.1 & 82.7  \\
15 & \bf 83.5 & 83.0 & 82.1 & 82.0 & 82.0   \\
20 & 81.0 & 81.8 & 81.6 & 81.8 & 82.3   \\
 \bottomrule
 \end{tabular}
    \caption{Sensitivity of the number of router experts (\textit{$N_r$}) and prompt lengths (\textit{L}) on SuperGlue RTE for T5-Base.}
    \label{tab:sensitivity}
    \vspace{-7mm}
\end{table}
To investigate the optimal length and number of prompt experts, we present sensitivity analyses for \our across varying configurations,
as detailed in Tab.~\ref{tab:sensitivity}. We have two observations: \textit{First}, \our consistently demonstrates performance improvements in prompt tuning across diverse settings, although the optimal configuration of soft prompt length and number may vary depending on the specific task. \textit{Second}, memory overhead during training and inference is solely dependent on the prompt length and irrelevant to the number of experts, as we only tune one set of router experts at a time due to the sparse MoE design, distinguishing us from vanilla prompt tuning methods. This provides a strong foundation for scaling up the number of router experts to tackle more complex and diverse contexts.

\section{Analytic Findings}\label{sec:expert}

\begin{figure}[t]        
    \centering
    
    \includegraphics[width=0.48\textwidth]{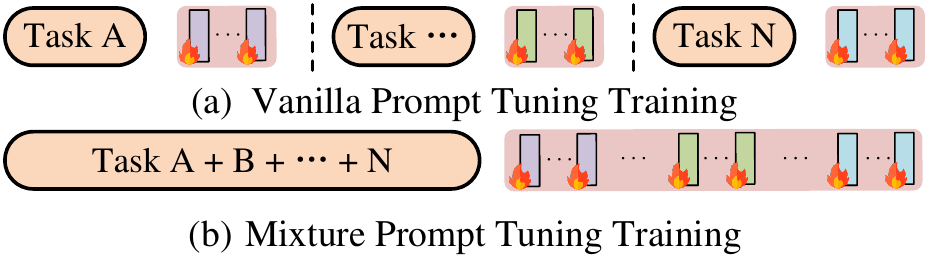}
    
    \caption{Overview of the two distinct training schemes employed in our study.}
    
    \label{fig:scheme}
    \vspace{-3mm}
\end{figure}

\noindent\textbf{Results on Mixture Training Scenario}.
As shown in Fig.~\ref{fig:scheme}, the vanilla prompt tuning training scheme trains separate prompts (\ie, different prompt colors represent distinct trained prompt embeddings) for each task, as distinct tasks require different prompt cues. In contrast, combining all subdatasets into a single dataset and training a unified prompt embedding for inference across all tasks imposes a higher demand on prompt generality (see Tab.~\ref{tab:combined}).

\begin{figure*}[ht]
  \centering
  \includegraphics[width=0.99\linewidth]{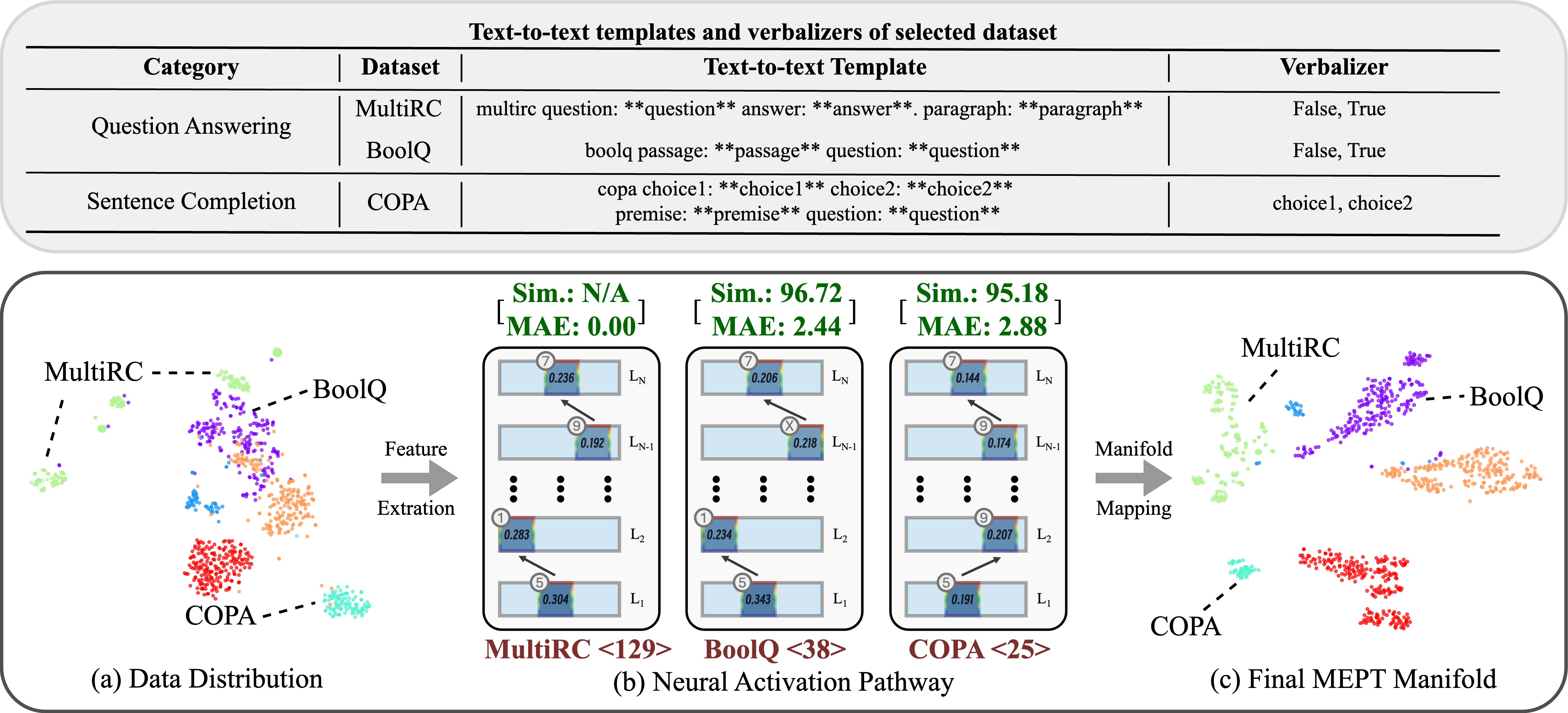}
  \vspace{-2mm}
  \caption{\textbf{Neural activation pathway for various tasks in the manifold mapping.} The attached table reveals the differences in SuperGLUE. (a) t-SNE visualizations of data distribution using BERT \cite{devlin2018bert}. (b) The neural activation pathway for each task, where the highest-probability expert in each layer is highlighted. The Mean Absolute Error (MAE) and Cosine Similarity (Sim.) with MultiRC are also computed respectively. The ``Semantic Similarity Score of Each Task with MultiRC'' is reported in <·>. (c) t-SNE visualizations of the final manifold representation from the SuperGlue development set in Tab.~\ref{tab:combined}. See more results and details in \S\ref{appendix:vis} and \S\ref{appendix:neural_pathway}.} 
  \label{fig:vis}
  \vspace{-3mm}
\end{figure*}

Compared to separate training, mixture training exhibits a significant performance degradation (\eg, 73.2 $vs.$ 76.9 average score for P-Tuning v2) due to the use of a single prompt across all tasks. Nevertheless, even under these constraints, \our consistently outperforms prior approaches through dynamic expert selection (see Fig.~\ref{fig:vis}), yielding a greater performance improvement (+1.1 $vs.$ +2.6).
\vspace{-2mm}

\noindent\textbf{Manifold Mapping Pathway}.
\begin{table}[t]
\begin{adjustbox}{width=0.85\width,center}
\small
    \centering
    \begin{tabular}{c|p{0.5cm}p{0.5cm}p{0.5cm}p{0.5cm}p{0.5cm}p{0.5cm}>{\columncolor[gray]{0.9}}c}
\toprule
Method & Boolq & CB & COPA &  MRC  &  RTE & WiC   &Ave. \\
\midrule
\midrule
\multicolumn{8}{c}{\textbf{Mixture Training on T5-Base} (220M)}\\
\midrule
Prompt-Tuning & 78.0 & 85.7 & 53.0 & 79.3 & 70.7   &63.7  & 71.7  \\
P-Tuning v2 & \underline{78.8} & \underline{85.7} & \bf 60.0 & 79.5 & 70.0  &\underline{65.0}& 73.2 \\
SMoP        & 78.6&85.7&58.0&\underline{79.7}&\underline{71.1}&64.8&73.0 \\
\our (Ours)   &\bf 80.1 &\bf 91.0& \underline{56.0}&\bf 80.2&\bf 79.7&\bf 67.8&  \bf 75.8   \\
 \bottomrule
 \end{tabular}
 \end{adjustbox}
    \caption{Performance comparison under mixture training scheme for T5-Base. See more results in Tab.~\ref{app:mixture}.}
    \label{tab:combined}
    \vspace{-5mm}
\end{table}
To the best of our knowledge, research on the understanding of prompt tuning remains rare~\cite{han2024facing, zeng2024visual}. 
Consequently, our research seeks to both quantitatively and qualitatively examine the impact of neural activation pathway on the enhancement of manifold mapping. 
Leveraging our intuitive MoE architecture, we can effectively visualize and analyze the activated expert that has the highest routing probability in each layer of \our, thereby constructing a pathway tailored to each task manifold mapping.
There are two key observations supporting the enhanced generality of \our.

\textit{Observation I: Neural activation pathways vary across distinct manifolds}. Observations from three different tasks neural pathways (\ie, MultiRC, BoolQ and COPA) in Fig.~\ref{fig:vis}(b) reveal that Neural activation pathways vary across distinct manifolds, reflecting task-specific requirements. Similar tasks (\ie, with higher semantic similarity score in <·>) tend to loyally activate overlapping pathways due to shared knowledge and the engagement of similar experts weight. For instance, the question answering datasets MultiRC and BoolQ demonstrate lower MAE (\ie, 2.88 $vs.$ 2.44) and higher cosine similarity (\ie, 95.18 $vs.$ 96.72) compared to COPA, a sentence completion task. More specifically, MultiRC and BoolQ tend to exhibit greater similarity in activated weights than COPA (\ie, 0.343 $vs.$ 0.191 in $L_1$) and are more likely to activate the same expert in each layer (\ie, MultiRC and BoolQ both activate expert 1 in $L_2$, while COPA prefers expert 9).
This observation highlights the heterogeneous learning preferences and capabilities of the experts.

\textit{Observation II: The improved linear separability of manifold mappings is facilitated by MEPT}. We observe significantly improved separability in manifold mappings in the final model outputs, as illustrated in Fig.~\ref{fig:vis}(a) and Fig.~\ref{fig:vis}(c). Specifically, the data distribution of combined dataset exhibits overlapping and poorly separable task representations (\ie, orange and purple groups tend to entangle closely together, whereas the green group is scattered throughout the space). In contrast, training with \our facilitates linear separability of manifolds (\ie, each color group maintains a distinct margin across different classes) and extraordinary manifold mappings (\eg, green group forms a more compact cluster). This separability aligns with our superior performance presented earlier in Tab.~\ref{tab:combined}.

\section{Conclusion}
We present \textbf{M}ixture of \textbf{E}xpert \textbf{P}rompt \textbf{T}uning, a manifold mapper that leverages the MoE framework to flexibly activate different neural pathways, resulting in a dynamic and general fine-tuning. It has merits in: 
\textbf{i)} demonstrating generality across various model sizes through expert prompts learning; \textbf{ii)} further compacting the number of parameters, thereby improving scalability; \textbf{iii)} thoroughly investigating the essence of manifold learning to elucidate the extraordinary MoE architecture.
The outcomes impart essential understandings of MoE design from a novel manifold mapper perspective, facilitating further exploration within this realm.

\clearpage
\section{Acknowledgements}
This research was supported by the National Science Foundation under Grant No. 2242243 \& 2348468.
\section*{Limitations}
\label{appendix:discussion}
For potential limitations, \our introduces an additional hyper-parameter—the number of router experts (\ie, $N_r$ in \S\ref{method:ept}). However, due to the sparse MoE structure (see analysis in \S\ref{sec:abla}), scaling up the number of router experts incurs minimal additional training and inference overhead. Furthermore, in practical applications, we observe in \S\ref{sec:abla} that, in general, a larger number of router experts tends to yield greater performance improvements, which serves as a guideline for selecting an appropriate number of router experts. Moreover, the efficiency of hyper-parameter selection can be further enhanced by incorporating a lightweight automatic expert number search mechanism.

Regarding the performance degradation under the mixture training scheme, it may stem from the imbalance in the dataset across the six SuperGLUE tasks (see Tab. \ref{app:dataset}), leading to unequal updates of expert prompts and consequently hindering the specialization of certain expert subsets. We plan to further explore its potential on larger and more diverse NLU tasks.

\bibliography{reference}

\clearpage
\appendix
\centerline{\maketitle{\textbf{SUMMARY OF THE APPENDIX}}}
This appendix contains additional details for the 63rd Annual Meeting of the Association for Computational Linguistics submission, titled \textit{``Mixture of Expert Prompt Tuning as a Manifold Mapper''}. The appendix is organized as follows:
\begin{itemize}
    \item \S\ref{appendix:dataset} provides an additional \textbf{introduction of the datasets} used, including the number of examples and task categories.
    \item \S\ref{appendix:details} explains \textbf{more implementation details} on backbone settings, baseline settings, training time, tokenization, and prepocessing.
    \item \S\ref{appendix:peft} presents further \textbf{performance comparison with other PEFT methods}.
    \item \S\ref{appendix:vis} includes supplemented \textbf{t-SNE, metric calculation and visualization details} in \S\ref{sec:expert}. Overall \textbf{expert utilization} is also provided and discussed.
    \item \S\ref{appendix:case} incorporates \textbf{extra ablative and pilot studies}.
    \item \S\ref{appendix:caseshared} provides two representative \textbf{case studies} for understanding the impact of shared experts.
    \item \S\ref{appendix:dis} provides discussions on \textbf{licenses, social impact, potential risks and directions of our future work}.
    \item \S\ref{appendix:neural_pathway} further includes more \textbf{detailed neural pathway visualizations and analysis}.
\end{itemize}

\section{Dataset Statistics}
\vspace{-2mm}
\label{appendix:dataset}
\begin{table}[h]
\begin{adjustbox}{width=1\columnwidth,center}
\begin{tabular}{ccccc}
\toprule
\bf Dataset & \bf Examples &\bf Task & \bf Domain & \bf Metric \\
\midrule
Boolq & 9,427 & QA & Wikipedia & Acc \\
CB & 250 & NLI & various & F1/Acc \\
COPA & 400 & SC & blogs, encyclop & Acc \\
MRC & 27,243 & QA & various & F1a \\
RTE & 2,490 & NLI & news, Wiki & Acc \\
WiC & 5,428 & WSD & lexical databases & Acc \\
\bottomrule
\end{tabular}
\end{adjustbox}
\caption{The details of 6 SuperGLUE tasks used in our
experiments. Datasets are classified into different task categories by~\cite{T5}. NLI denotes natural language inference, QA denotes questions and answers task, SC denotes sentence completion, WSD denotes word sense disambiguation, Acc denotes accuracy, F1a denotes the macro F1 score. }\label{app:dataset}
\vspace{-5mm}
\end{table}
Tab. \ref{app:dataset} shows details of the six datasets from
SuperGLUE benchmark \cite{SuperGLUE} that we
used for our experiments, along with their training
sizes and evaluation metrics. Following \cite{T5} and \cite{lester2021power}, for tasks that have two evaluation metrics we use the average of both scores as the final performance metric.

\section{Training Details}
\label{appendix:details}
\subsection{Backbone Settings}
This paper employs two distinct model architectures (\ie, decoder-only and encoder-decoder) and three different backbone sizes: T5-Base, T5-Large, and Llama3.2, with parameter counts of 200M, 770M, and 1B, respectively.

The performance gap between LLaMA 3.2 and T5 on SuperGLUE under prompt tuning stems from differences in pretraining objectives and architecture. LLaMA, trained with causal language modeling, excels at autoregressive generation but struggles with structured NLU tasks due to its mask strategy and decoder-only architecture, which limits bidirectional context integration. In contrast, T5’s span-corruption objective fosters bidirectional reasoning, and its encoder-decoder architecture captures richer contextual dependencies. Notably, prior prompt tuning methods \cite{aprompt, liu2022few} for T5 have exclusively prepended prompts to the encoder layer while leaving the decoder layer unchanged, as the latter must adapt to various downstream tasks. In our study, we explored extending prompt embeddings to both the encoder and decoder layers. However, preliminary results revealed significant performance degradation, corroborating the poor outcomes observed in LLaMA, a decoder-only model.
These factors make T5 inherently more suited for prompt tuning, as its training paradigm aligns well with structured task formulations.

\subsection{Baseline Settings}

\noindent\textbf{Fine-Tuning} \cite{ExT5} refers to the standard full fine-tuning approach for T5, wherein all parameters are updated during training.

\noindent\textbf{Prompt Tuning} \cite{lester2021power,liu2025reimaginingmultimodalinstructiontuning} is a foundational approach that introduces trainable prompts at the first input layer. XPrompt \cite{xprompt} enhances efficiency by pruning less informative token-level and piece-level prompts. P-Tuning v2 \cite{ptuningv2} extends Prompt-Tuning by injecting distinct prompt embeddings into each Transformer layer. ResPrompt \cite{ResidualPrompt} further improves performance and stability by incorporating residual connections into the prompt tuning framework. DePT \cite{shi2023dept} decomposes the soft prompt into a shorter soft prompt and a pair of low-rank matrices, which are optimized using two distinct learning rates. SuperPos-Prompt \cite{pmlr-v262-ali-sadraei-javaheri24a} introduces a novel reparameterization technique that leverages the superposition of multiple pretrained vocabulary embeddings to enhance soft prompt learning. SMoP \cite{choi-etal-2023-smop} utilizes a gating mechanism in the first layer to train multiple short soft prompts, each specialized for different data subsets, offering an alternative to using a single long soft prompt for the entire dataset. EPT \cite{lan2024efficient} leverages multi-space projection and prompt fusion. Specifically, it decomposes a given soft prompt into a shorter prompt and two low-rank matrices, optimizing them to enhance efficiency and performance. Notably, we linearly search the best learning rate from \{1e-5, 5e-5, 1e-4\} for MEPT and \{0.1, 0.3, 0.5\} for other prompt tuning methods following \cite{choi-etal-2023-smop}.

\noindent\textbf{Linear} serves as the final tunable score network in LlamaForSequenceClassification and has been evaluated exclusively within the Llama backbone framework. Notably, in all prompt tuning methods based on the Llama backbone, both the prompt and the final score network are fine-tuning.
\subsection{Tokenization 
Preprocessing}
Following common practice \cite{lester2021power}, we set the maximum input length, including the prompt, to 512 tokens for all experiments. Inputs are padded to this length, with padded tokens masked out, and truncated if they exceed 512 tokens. No additional preprocessing (e.g., punctuation removal) is applied; instead, raw text from SuperGLUE datasets is directly tokenized using the respective model’s tokenizer. All experiments adhere to the SMoP \cite{choi-etal-2023-smop} formatting, where classification tasks are reformulated into text-to-text format in T5 model. For instance, in BoolQ, labels ‘0’ and ‘1’ are mapped to ‘True’ and ‘False,’ respectively. In T5, we translate each SuperGLUE dataset into a text-to-text format following ~\cite{choi-etal-2023-smop}. In Llama, we continue to use the previously established text-to-text template while preserving the original labels, aligning the task with LlamaForSequenceClassification.

\subsection{Router Details}

\begin{itemize}
    \item \textbf{Stochastic:} Use \texttt{torch.randint} to randomly select a router expert instead of always choosing the top-1 expert.
    
    \item \textbf{Dense:} Use all router experts simultaneously rather than selecting only the top-1.
    
    \item \textbf{Gumbel-Softmax:} Add Gumbel noise (temperature set to 1) to the logits, then apply a softmax function before selecting the router expert.
    
    \item \textbf{Perturbation:} Add Gaussian noise to the router with standard deviation $\sigma = 1$.
\end{itemize}

\subsection{Text-to-text templates and verbalizers of SuperGLUE}

BoolQ
\begin{graybox}
\textbf{Template:} boolq passage: **passage** question: **question**
\noindent \textbf{Verbalizer:} False, True
\end{graybox}

CB
\begin{graybox}
\textbf{Template:} cb hypothesis: **hypothesis**. premise: **premise**
\noindent \textbf{Verbalizer:}
entailment, contradiction, neutral
\end{graybox}

COPA
\begin{graybox}
\textbf{Template:} copa choice1: **choice1** choice2: **choice2** premise: **premise** question: **question**
\noindent \textbf{Verbalizer:}
choice1, choice2
\end{graybox}

MultiRC
\begin{graybox}
\textbf{Template:} multirc question: **question** answer: **answer**. paragraph: **paragraph**
\noindent \textbf{Verbalizer:}
False, True
\end{graybox}

RTE
\begin{graybox}
\textbf{Template:} rte sentence1: **premise** sentence2: **hypothesis**
\noindent \textbf{Verbalizer:}
entailment, not\_entailment
\end{graybox}

WiC
\begin{graybox}
\textbf{Template:} wic sentence1: **sentence1** sentence2: **sentence2** word: **word**
\noindent \textbf{Verbalizer:}
False, True
\end{graybox}

\subsection{Training Time}

\begin{table}[h]
\begin{adjustbox}{width=1\columnwidth,center}
\begin{tabular}{cccccccc}
\toprule
Methods & \textbf{Boolq} & \textbf{CB} & \textbf{COPA}  &   \textbf{RTE} & \textbf{WiC}   \\
\midrule
Prompt-Tuning$^\ast$& 	2h38m& 	8m& 11m&  57m& 51m  \\
P-Tuning v2$^\ast$& 3h36m&	10m&	14m&		1h28m&	1h14m\\
XPrompt$^\ast$&	4h47m&	14m	&29m&		2h26m&	2h19m\\
ResPrompt$^\ast$&	3h47m&	10m	&23m&		1h51m&	1h34m\\
\our & 3h23m &6m&8m&50m&35m\\
\bottomrule
\end{tabular}
\end{adjustbox}
\caption{Training time of \our \ on SuperGLUE. `$^\ast$' indicates the results reported from \cite{aprompt}.}\label{app:time}
\vspace{-3mm}
\end{table}
We further analyze and report the training time of different methods across all SuperGLUE tasks in T5-Base in Tab. \ref{app:time}. In Prompt-Tuning, the trainable parameters are limited to the prompts introduced at the input layer. For P-Tuning V2, trainable parameters extend across all layers, encompassing prompts throughout the model. XPrompt, following a pruning process, retains only a subset of prompts as trainable parameters, specifically at the input layer. In contrast, ResPrompt includes both input prompts and additional residual network components as trainable parameters. The total number of trainable parameters for each approach is summarized in Tab. \ref{overall_comparison}. Due to the sparsity inherent in the MoE design, \our\ benefits from a relatively shorter training time, as only a subset of prompts is updated in each iteration (\ie, 10 prompts in \our \textit{vs.} 100 prompts in P-Tuning v2).

\section{Comparison with Other Parameter Efficient Methods}\label{appendix:peft}
\vspace{-2mm}
\begin{table}[h]
\begin{adjustbox}{width=1\columnwidth,center}
\begin{tabular}{c|cccc}
\toprule
\bf Methods & \bf Boolq & \bf CB & \bf RTE & \bf WiC \\
\midrule
Fine-Tuning \cite{ExT5}                          & 85.75 & 95.26 & 88.05  & 72.11 \\
Adapter$^*$ \cite{DBLP:conf/emnlp/PfeifferRPKVRCG20} & 85.30 & 92.30 & 87.30  & 69.80 \\
LoRA$^*$ \cite{hu2021lora}	                         & 85.10 & 92.60 & 87.30  & 70.20 \\
\our (Ours) & \bf85.96   & \bf 97.61  & \bf   90.01 & \bf 73.09  \\
\bottomrule
\end{tabular}
\end{adjustbox}
\caption{Performance comparison with other non-prompt tuning based parameter efficient methods on Boolq, CB, RTE and WiC for T5-Large. `$^\ast$' indicates the results reported from \cite{jin2023parameter}.}\label{other_pem}
\vspace{-4mm}
\end{table}

\label{appendix:compare_pem}
To comprehensively evaluate performance, we compare our approach against other parameter-efficient methods that do not rely on prompt tuning, including Adapter \cite{DBLP:conf/emnlp/PfeifferRPKVRCG20} and LoRA \cite{hu2021lora}. More recent LoRA-based and adapter-based methods have been developed on different backbones and datasets, but their unpublished code makes reproduction under the same settings infeasible. The performance comparison results are presented in Tab. \ref{other_pem}, showing that \our surpasses other parameter-efficient baselines by a significant margin. Notably, existing prompt tuning approaches also demonstrate superior performance compared to these methods.

\section{Visualization Details}
\label{appendix:vis}
\noindent\textbf{Manifold Analysis Details}. 
To analyze the learned feature representations across different tasks, we extract features from the encoder's last hidden layer by averaging over the sequence length dimension. For balanced comparison, we sample a maximum of 200 examples from each dataset's validation set. The features are then projected into a 2D space using t-SNE with a perplexity of 30 and random seed 42. We average the sequence-level representations ($h \in \mathbb{R}^{L \times d}$ where L is the sequence length and d is the hidden dimension) to obtain a fixed-size vector ($\bar{h} \in \mathbb{R}^d$) for each input, enabling direct comparison across examples of varying lengths. These embeddings are then visualized in a 2D scatter plot where points are colored according to their source dataset, allowing us to examine the natural clustering of different tasks in the model's learned representation manifold.

\noindent\textbf{Metric Details}.
For analyzing model behavior, we employ different similarity metrics. To compare semantic similarities between different tasks, we utilize BERT Similarity Score which computes the dot product of L2-normalized BERT features ($BERTSim(b_1, b_2) = \frac{b_1}{||b_1||_2} \cdot \frac{b_2}{||b_2||_2}$). For neural pathway analysis to compare expert routing patterns, we use two complementary metrics: Mean Absolute Error (MAE) which measures the average magnitude of differences between expert utilization patterns ($MAE(x, y) = \frac{1}{n}\sum_{i=1}^{n} |x_i - y_i|$), and Cosine Similarity which quantifies the directional similarity between patterns ($CosSim(x, y) = \frac{x \cdot y}{||x|| \cdot ||y||}$).

\begin{figure}[h]        
    \centering

    \includegraphics[width=0.45\textwidth]{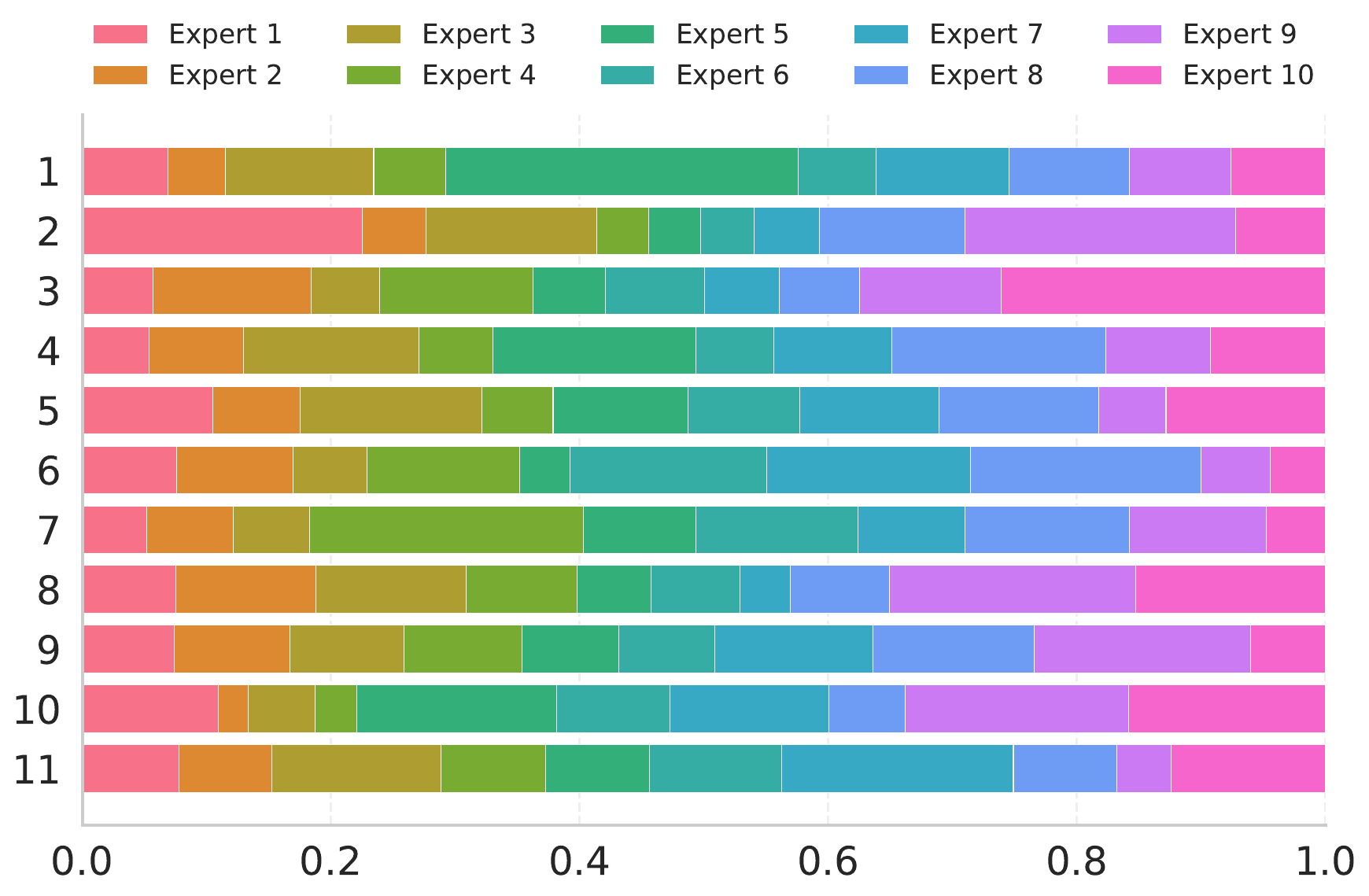}
    \caption{The expert utilization in each layer under the mixture training scheme.}
    \label{fig:expert_uti}
    \vspace{-3mm}
\end{figure}

\noindent\textbf{Expert utilization}.
We present a visualization of the expert utilization across six tasks under the mixture training in Fig.~\ref{fig:expert_uti}. For each layer, the bar length in different colors corresponds to the activated frequency assigned to individual experts. As the total expert utilization are normalized to sum to 1, the total bar length for each layer remains constant. At a macro level, the visualization clearly illustrates significant variation in the contributions of each expert (\ie, experts do not contribute equally across all layers), reinforcing the notion that different experts are specialized in distinct facets of natural language processing (\ie, different layers exhibit preferences for distinct experts). 

\section{Ablative and Pilot Studies}
\label{appendix:case}
\vspace{-2mm}
\begin{table}[h]
\small
    \centering
    \begin{tabular}{l| ccccc}
\toprule
\multicolumn{1}{c}{\our} &  CB & COPA  & WiC & MRC \\\midrule
Xavier Normal &90.9& 62.3& 69.6&\bf  80.7   \\
He Normal & 90.7 &62.7&69.0& 80.2 \\\midrule
Previous Knowledge &\bf 91.0& \bf 63.7 &\bf 69.9&80.4  \\
 \bottomrule
 \end{tabular}
    \caption{Performance comparison with different prompt
initializations on CB, COPA, WiC and MRC for T5-Base.
    }
    \label{tab:main_ablation}
    \vspace{-3mm}
\end{table}

\noindent\textbf{Prompt Initialization.} As parameter initialization significantly influences the learning process~\cite{glorot2010understanding}, we further conduct experiments under three different initialization strategies. These include commonly used methods such as Xavier normal~\cite{glorot2010understanding} and He normal~\cite{he2015delving}, and a previous-knowledge-based initialization that leverages stored parameters from prior layer training to facilitate a quicker start. Our results reveal that 
traditional initialization methods are stable in \our, with 
Xavier normal showing a slight advantage across various tasks. Notably, the previous-knowledge-based initialization demonstrates the best performance, consistent with the intuition that leveraging prior training can provide better local minima. However, this approach requires a two-stage training process, doubling the training time and rendering it less cost-effective for practical applications. As a result, we apply
xavier normal as the default initialization strategy in our framework. \\

\label{appendix:mixture}
\begin{table}[h]
\vspace{-2mm}
\begin{adjustbox}{width=1\columnwidth,center}
\begin{tabular}{ccccccccc}
\toprule
\bf \# & \bf CB  & \bf COPA & \bf WiC & \bf RTE & \bf  Boolq & \bf MRC & \bf Gain\\
\midrule
1 &\checkmark &&&&&& +0.7\\
2 &\checkmark&\checkmark&&&&& +1.0\\
3 &\checkmark&\checkmark&\checkmark&&&&+1.3 \\
4 &\checkmark&\checkmark&\checkmark&\checkmark&&&+1.6 \\
5 &\checkmark&\checkmark&\checkmark&\checkmark&\checkmark&& +2.1\\
6 &\checkmark&\checkmark&\checkmark&\checkmark&\checkmark&\checkmark& +2.6\\
\bottomrule
\end{tabular}
\end{adjustbox}
\caption{Performance comparison under different mixture training settings. ``Gain'' denotes the absolute performance improvement over the previous best prompt tuning method, P-Tuning v2.}\label{app:mixture}
\vspace{-3mm}
\end{table}

\noindent\textbf{Mixture Training.} To further investigate the performance of \our under different mixture training settings, we additionally experiment with combining 2 to 5 datasets (\ie, settings \#2 to \#5) for training. A greater number of combined datasets introduces higher diversity, thereby increasing the requirement for prompt generalization. In the non-mixture training setting \#1, \our demonstrates the lowest performance gain, potentially due to the relatively homogeneous distribution within CB, which leads the MoE-based prompt tuning to select nearly identical neural pathways. As the number of combined datasets increases, performance improvements become more pronounced (\ie, from +1.0 to +2.6), highlighting the potential of \our for handling more complex and diverse large-scale datasets.

\vspace{-2mm}
\begin{table}[h]
\small
    \centering
    \begin{tabular}{l| cccccc}
\toprule
 &  1 & 1→3  & 1→6 & 1→9 & 1→12 \\\midrule
CB &86.4& 88.5& 89.8&89.9 & \bf 90.9  \\
COPA & 	58.3 &60.7&61.3& 62.0 & \bf 62.3
\\
 \bottomrule
 \end{tabular}
    \caption{Additional ablation on prompt depth. $i$ → $j$ indicates the layer indices where prompts are inserted, with the 1st layer being closest to the input. T5-base consists of 12 layers in total.
    }
    \label{appendix:depth}
    \vspace{-3mm}
\end{table}

\noindent\textbf{Additional Ablation on Prompt Depth.} we further conduct additional prompt depth experiments on CB and COPA and present the results in Tab.~\ref{appendix:depth}. Our findings reveal two key observations. First, we observe a consistent performance improvement as depth increases, demonstrating that deeper MoE prompts indeed contribute to learning better representations. Second, the performance gain is most significant when increasing the depth from 1 to 1-3 (\eg, +2.4 on COPA), while the improvement becomes marginal from 1-9 to 1-12. This saturation trend aligns with previous research~\cite{jia2022visual}, indicating that prompt depth beyond a certain point yields diminishing returns.

\vspace{-2mm}
\begin{table}[h]
\small
    \centering
    \begin{tabular}{l| cccccc}
\toprule
 &  4:0 & 4:1  & 4:2 & 4:3 & 4:4 \\\midrule
RTE &82.3	&83.5&	83.3&83.5&\bf	83.6  &
\\
 \bottomrule
 \end{tabular}
    \caption{Ablation Study on Routed-to-Shared Expert Ratio. Each configuration is denoted as $a{:}b$, representing the ratio between the routed experts ($a$) and the shared experts ($b$).
    }
    \label{tab:expert-ratio}
    \vspace{-3mm}
\end{table}

\noindent\textbf{Routed-to-Shared Expert Ratio.} To further investigate the fine-grained impact of the expert ratio, we conducted an additional experiment on the RTE dataset using T5-Base. The results are summarized in Tab.~\ref{tab:expert-ratio}. As shown, increasing the number of shared experts leads to only marginal performance improvements, which is consistent with previous research~\cite{liu2024deepseek}. We hypothesize that this is primarily due to the relatively small scale and simplicity of the RTE task. In future work, we plan to extend this analysis to more complex and challenging datasets..

\begin{table}[h]
\small
\centering
\begin{tabular}{l|ccc}
\toprule
\multicolumn{1}{c}{} & T5-3B & LLaMA2-7B & T5-11B \\
\midrule
Prompt Tuning & 82.6 & 84.1 & 83.3 \\
DePT          & 84.0 & 84.8 & 84.5 \\
EPT           & 85.5 & 85.3 & 85.9 \\
\textbf{MEPT} & \textbf{86.6} & \textbf{86.5} & \textbf{87.0} \\
\bottomrule
\end{tabular}
\caption{Preliminary RTE results (\%) with quantized large foundation models.}
\label{tab:prelim_rte}
\vspace{-3mm}
\end{table}

\noindent\textbf{Preliminary Results on Large Foundation Models}. We explore the potential of prompt tuning on larger foundation models and have conducted preliminary experiments on the RTE task using T5-3B, LLaMA2-7B, and T5-11B. To reduce GPU memory usage, we applied quantization to these models. MEPT consistently achieves superior performance. 

\begin{table}[h]
\small
\centering
\begin{tabular}{l|cc}
\toprule
\multicolumn{1}{c}{} & SQuAD 1.1 Dev & SQuAD 2.0 Dev \\
\midrule
Fine-Tuning  & \textbf{91.0} & \textbf{82.0} \\
P-Tuning v2  & 89.5 & 77.8 \\
MEPT         & 90.2 & 78.6 \\
\bottomrule
\end{tabular}
\caption{Results on SQuAD 1.1 and 2.0 development sets. MEPT is competitive with prompt-tuning baselines but lags behind full fine-tuning.}
\label{tab:squad_results}
\vspace{-3mm}
\end{table}

\noindent\textbf{Extension of MEPT to Question Answering}. We further extend MEPT to more diverse task settings. Specifically, we apply MEPT to the SQuAD dataset~\cite{SQuAD}, including both versions 1.1 and 2.0. This requires the model to parse passages and answer questions with a higher level of comprehension, beyond classification-style tasks. As shown in  Tab.~\ref{tab:squad_results}, MEPT consistently achieves competitive performance. However, both prompt tuning methods underperform compared to full fine-tuning, highlighting the limitations of prompt tuning for more complex and fine-grained tasks.

\noindent\textbf{Mixture Training for Cross-Task Generalization}. We include mixture training settings to evaluate prompt generality. We further conducted experiments by training on MultiRC and testing on BoolQ using T5-Base during rebuttal. As shown in Tab.~\ref{tab:mixture}, MEPT achieves notably better cross-task generalization.

\begin{table}[h]
\small
\centering
\begin{tabular}{l|ccc}
\toprule
\multicolumn{1}{c}{} 
& \shortstack{BoolQ \\ $\downarrow$ \\ BoolQ} 
& \shortstack{MRC \\ $\downarrow$ \\ BoolQ} 
& \shortstack{All \\ $\downarrow$ \\ BoolQ} \\
\midrule
Prompt-Tuning & 78.1 & 78.0 & 78.0 \\
P-Tuning v2   & 80.8 & 79.4 & 78.8 \\
\textbf{MEPT} & \textbf{81.1} & \textbf{80.8} & \textbf{80.1} \\
\bottomrule
\end{tabular}
\caption{Results of cross-task generalization on BoolQ using T5-Base. 
$A \to B$ indicates training on dataset $A$ and evaluating on dataset $B$. 
“All” refers to training on all six SuperGLUE datasets (see Tab.~\ref{tab:combined} for additional results).}
\label{tab:mixture}
\vspace{-3mm}
\end{table}

\section{Case Study}
\label{appendix:caseshared}

Our design of shared experts is inspired by DeepSeek~\cite{liu2024deepseek}, which emphasizes capturing common knowledge and minimizing redundancy among routed experts. In our study, we find that within the SuperGLUE benchmark, such ``common knowledge'' is often aligned with semantic understanding across tasks. To illustrate this, we provide two representative case studies:

\noindent\textbf{BoolQ (Boolean Questions).}  
BoolQ is a classic question-answering dataset grounded in general knowledge. In the example below, the model is asked whether Iran and Afghanistan share the same language based on a provided passage. Without shared experts, the model fails to answer correctly (predicting 0 instead of the ground-truth label 1). This failure suggests that shared experts play a key role in encoding factual knowledge that spans across inputs. 

\begin{quote}
\textit{Question:} do iran and afghanistan speak the same language \\
\textit{Passage:} Persian language -- Persian, also known by its endonym Farsi, is one of the Western Iranian languages within the Indo-Iranian branch of the Indo-European language family. It is primarily spoken in Iran, Afghanistan (officially known as Dari since 1958), and Tajikistan (officially known as Tajiki since the Soviet era), and some other regions which historically were Persianate societies and considered part of Greater Iran. It is written in the Persian alphabet, a modified variant of the Arabic script, which itself evolved from the Aramaic alphabet. \\
\textit{Label:} 1
\end{quote}

\noindent\textbf{WiC (Word-in-Context).}  
WiC is designed to evaluate models' ability to capture context-sensitive word meanings. In the following case, the word ``place'' carries different meanings: a physical location in the first sentence and a metaphorical position in the second. Without shared experts, the model incorrectly predicts the meanings to be the same, indicating its failure to distinguish subtle semantic nuances. This supports the role of shared prompts in improving semantic representation learning.

\begin{quote}
\textit{Word:} place \\
\textit{Sentence 1:} Do you want to come over to my place later? \\
\textit{Sentence 2:} A political system with no place for the less prominent groups. \\
\textit{Label:} 0
\end{quote}

These qualitative findings are consistent with the quantitative results presented in Tab.~\ref{tab:router}, where the absence of shared experts leads to a noticeable performance drop on WiC (from 69.6 to 67.0). Together, these observations underscore the importance of shared experts in enhancing semantic understanding across tasks.

\begin{figure*}[h]
  \centering
  \includegraphics[width=0.99\linewidth]{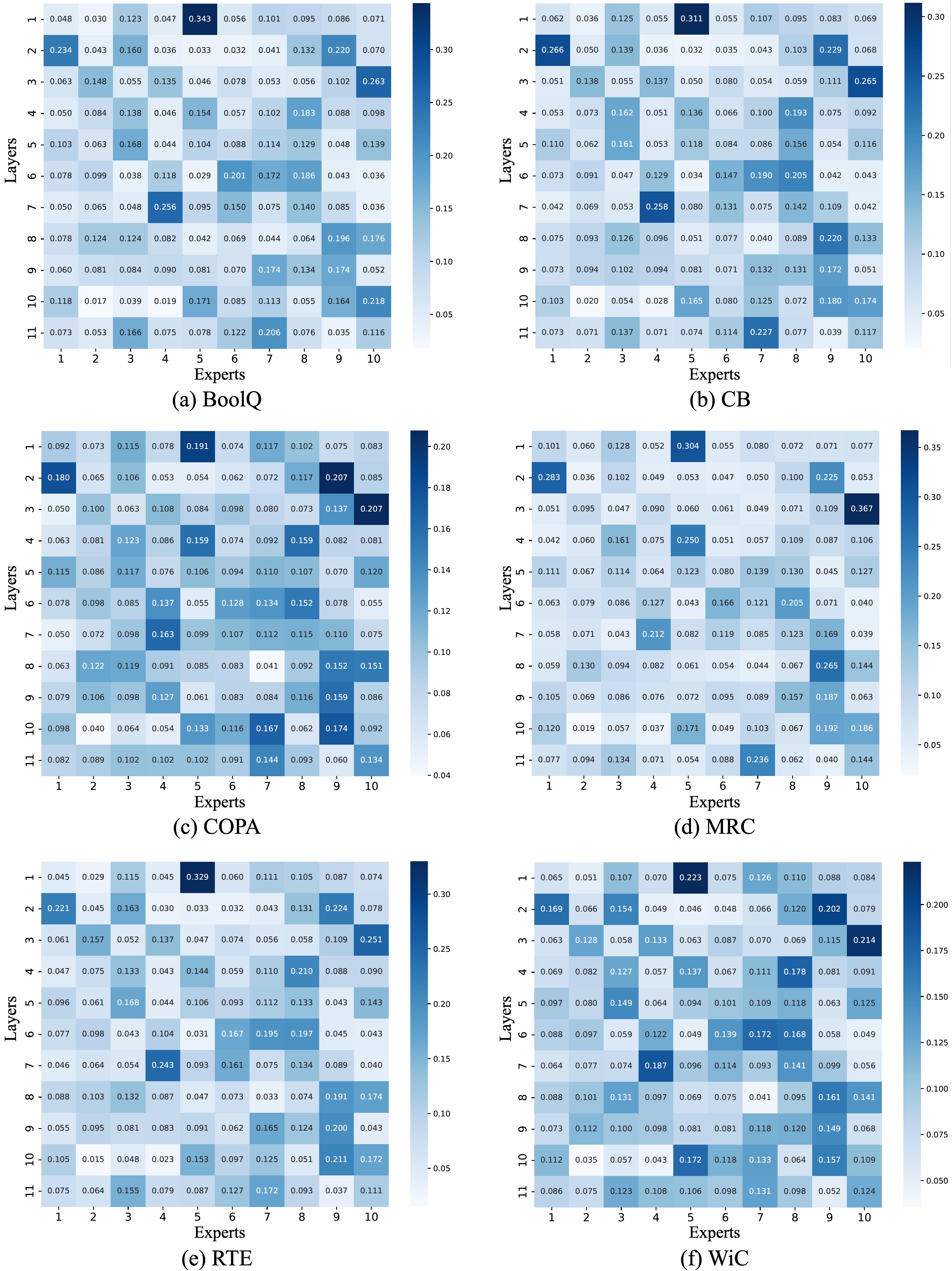}
  \vspace{-2mm}
  \caption{\textbf{Detailed neural activation pathway for six SuperGLUE tasks.} Each row is normalized to sum to 1, with the activated experts in each layer represented by the deepest color, indicating the probability of selecting a given expert.} 
  
  \label{fig:neural_p}
  \vspace{-6mm}
\end{figure*}
\vspace{-2mm}
\section{Discussion}
\label{appendix:dis}
\vspace{-1mm}
\subsection{Manifold View of Layer-wise Abstraction}

We analyze the network through the lens of manifold learning~\cite{Papyan_2020,kirsanov2025geometrypromptingunveilingdistinct}. As depth increases, transformer representations become more abstract and discard nuisance variation irrelevant to the task.

For a network with $L$ layers, each layer $l$ transforms the representation as follows:
$$
h^{(l)} = f\Bigl(h^{(l-1)}\Bigr)
$$

The abstraction process can be illustrated by:
$$
x \to h^{(1)} \to h^{(2)} \to \dots \to s 
$$
$$
(\text{from concrete to abstract})
$$

The representation space progressively collapses around class-relevant information:
$$
\dim\Bigl(\mathrm{Var}(h^{(l)} \mid y)\Bigr) \downarrow \quad \text{as } l \uparrow
$$

This collapse exhibits several properties:

\textbf{1. Within-Class Convergence:} As $l$ increases,
$$
\|\|h^{(l)}(x_1) - h^{(l)}(x_2)\|\| \to 0 
$$
$$
\text{for } x_1, x_2 \text{ in the same class}
$$

\textbf{2. Between-Class Separation:} Simultaneously,
$$
\|\|h^{(l)}(x_1) - h^{(l)}(x_2)\|\| \to d > 0 
$$
$$
\text{for } x_1, x_2 \text{ in different classes}
$$

\textbf{3. Final Collapse:} At the top layer,
$$
h^{(L-1)}(x) \approx v_c \quad \text{for all } x \text{ in class } c,
$$
where $v_c$ is the prototype vector for class $c$.

These phenomena motivate our method. Fig.~\ref{fig:vis} shows within-class convergence (\eg, the green cluster tightens), between-class separation (distinct inter-class margins), and final collapse (compare Fig.~\ref{fig:vis}(a) and Fig.~\ref{fig:vis}(c)).

\subsection{Key Distinctions from SMoP}
To highlight the advantages of MEPT, we summarize its key distinctions from SMoP~\cite{choi-etal-2023-smop} as follows. First, MEPT employs a deep multi-layer MoE architecture, incorporating multiple experts at every Transformer layer, while SMoP restricts MoE application to the input layer, thereby limiting its representational capacity. Second, MEPT introduces a hybrid design of shared and non-shared experts, enabling it to capture both generalizable knowledge and instance-specific features. In contrast, SMoP utilizes only non-shared experts, adhering to a traditional design. Third, MEPT integrates manifold learning with MoE through visualization, providing deeper insights into the internal dynamics of MoE-based prompt tuning.

\vspace{-1mm}
\subsection{Key Distinctions from DeepSeek MoE}\label{keydis}
Our design of routed and shared experts is inspired by ~\cite{liu2024deepseek,valipour2024sortednetscalablegeneralizedframework,hu2024amortizingintractableinferencelarge,bjerke2023understandingneuralcodinglatent,cohen2020separability}. However, there are two major differences between MEPT and the MoE approach used in DeepSeek~\cite{dai2024deepseekmoe}: 

\begin{itemize}
    \item \textit{MoE Placement and Structure}: Unlike DeepSeek and other traditional MoE methods, which apply MoE to the FFN module, MEPT incorporates the MoE structure into soft prompts that dynamically select prompt experts to enhance task adaptability. 
    \item \textit{Efficiency and Scalability}: MEPT benefits from a more efficient representation, making it easier to scale. Specifically, its inherently compact parameter space follows a leaner design (\eg, the parameter count for the prompt-based method is $mh$, where $m$ and $h$ represents the prompt length and hidden layer dimension, respectively), allowing for more scalable expansion compared to DeepSeekMoE. In contrast, the parameter count for the MLP component in a transformer is approximately $8h^2$.
\end{itemize}

We further in detailed discuss the differences in MoE structures below:

\begin{itemize}

 \item \textit{Expert Design}. DeepSeekMoE introduces Fine-Grained Expert Segmentation to facilitate more focused knowledge distribution across experts, resulting in a structure of 1 shared expert and up to 63 routed experts (\ie, 1 + 63). In contrast, MEPT operates within a much smaller parameter space (\ie, $mh$ $vs.$  $8h^2$), and thus does not adopt fine-grained segmentation. Instead, we limit the number of router experts to between 10 and 20.
 \item \textit{Load Balancing}. DeepSeekMoE employs both Expert-Level Balance Loss and Device-Level Balance Loss to mitigate the risk of routing collapse. In MEPT, all training is performed on a single GPU, eliminating the need for Device-Level Balance Loss. However, we were inspired by DeepSeekMoE’s use of Expert-Level Balance Loss and explored whether similar strategies could help prevent the model from over-relying on a small subset of experts. As shown in Tab.~\ref{tab:router}, row six (Perturbation), we applied Gaussian noise to the router outputs.
 \item \textit{Router Selection}. Both DeepSeekMoE and MEPT adopt Top-K routing; however, they differ in activation density. DeepSeekMoE uses a denser activation with K=7, resulting in 1 shared expert plus 7 routed experts (\ie, 1+7 out of 63), whereas MEPT employs a much sparser activation with K=1, activating 1 shared expert and 1 routed expert (i.e., 1+1 out of 10–20). We also explored whether MEPT could benefit from denser activation. However, increasing the activation density (\eg, 1+all) did not significantly improve the overall performance.
 \end{itemize}

\vspace{-1mm}
\subsection{Asset License and Consent}

The majority of prompt tuning \citep{ptuningv2, lester2021power} and T5 \cite{ExT5}, are licensed under \href{https://www.apache.org/licenses/LICENSE-2.0}{Apache-2.0}; Llama-3.2 1B \cite{dubey2024llama} is licensed under \href{https://huggingface.co/meta-llama/Llama-3.2-1B/blob/main/LICENSE.txt}{Llama 3.2 Community License Agreement}; SuperGLUE is licensed under \href{https://opensource.org/license/mit/}{MIT};
\vspace{-1mm}
\subsection{Artifact Consistent With Intended Use}
Our work ensures that the use of existing artifacts aligns with their intended purpose when specified. For the artifacts we create, it remains compatible with the original access conditions. In particular, we ensure that derivatives of data accessed for research purposes are confined to research contexts.

\vspace{-1mm}
\subsection{Social Impact}

This work introduces \our, which demonstrates significant performance improvements over state-of-the-art baselines, as shown in Tab. \ref{overall_comparison}, while requiring substantially fewer tunable parameters for prompt tuning. Our approach enhances model accuracy and is particularly beneficial for parameter-sensitive training scenarios, such as prompt tuning on resource-constrained devices and rapid adaptation with limited computational overhead.
\vspace{-1mm}

\subsection{Future Work}
\noindent\textbf{Extending to More Backbones and Datasets}: Future research can investigate the generalizability of our method by applying it to a broader range of backbone architectures (\ie, Deepseek and Qwen) and datasets (\ie, MMLU and GSM-8k). Evaluating performance across diverse model families and domains will provide deeper insights into the robustness and adaptability of the proposed approach.

\noindent\textbf{Scaling to a Larger and More Fine-Grained Expert System}: Inspired by recent advancements such as DeepSeek, we aim to explore increasing the number of experts while incorporating more fine-grained specialization across layers. This could improve both efficiency and performance by dynamically selecting more targeted expert pathways for different types of inputs.

\noindent\textbf{Applying to Causal Language Modeling}: While our current study focuses on sequence classification or sequence-to-sequence models, extending the approach to causal language models presents an exciting avenue for research. Adapting MoE-based prompt tuning to autoregressive settings could unlock new capabilities for large-scale generative models.
\vspace{-1mm}
\subsection{Potential Risks}
Consider the tuning process of LLM, which has potential risks for energy usage. Finetuning requires additional computational power, leading to energy use and increased environmental impact. 
\vspace{-1mm}
\subsection{AI Disclosure}
We acknowledge the use of GPT-4o for grammar correction and sentence-level refinement. The model was employed to enhance clarity, coherence, and fluency while ensuring the original meaning and intent of the text remained unchanged.

\vspace{-2mm}
\section{Neural Pathways}
\label{appendix:neural_pathway}

\vspace{-2mm}
As shown in Fig. \ref{fig:neural_p}, we conduct a detailed analysis and visualization of expert weights across all layers and tasks, where higher probabilities are represented by deeper colors. From this visualization, we derive two key observations.

First, the neural pathways across all six tasks exhibit substantial similarity, with only a few key differences in expert selection. This is likely attributable to the intrinsic similarities present in natural language. Additionally, the relatively small dataset size for each task may contribute to this overlap. Utilizing larger and more diverse datasets could potentially enhance the differentiation of neural pathways, leading to more distinct expert specializations.

Second, tasks within the same category exhibit greater similarity in neural pathway activation compared to those from different categories, as discussed in \S\ref{sec:expert}. This is primarily reflected in the closer distribution of expert weights, suggesting that related tasks share common feature representations and utilize similar expert specializations.

\end{document}